\documentclass[letterpaper, 10 pt, conference]{ieeeconf}  % Comment this line out if you need a4paper

\IEEEoverridecommandlockouts                              % This command is only needed if 
\usepackage{graphics} % for pdf, bitmapped graphics files
\usepackage{epsfig} % for postscript graphics files
\usepackage{amsmath} % assumes amsmath package installed
\usepackage{amssymb}  % assumes amsmath package installed
\usepackage{multirow}	% multi row for table

\DeclareMathOperator{\sgn}{sgn}
\newcommand{\mat}[1]{\ensuremath{\begin{bmatrix}#1\end{bmatrix}}}	% matrix
\newcommand{\diag}[1]{\text{diag}(#1)}							% diag
\newcommand{\x}{\ensuremath{\times}}

\usepackage{amsmath,bm,times}
\usepackage[usenames,dvipsnames,table]{xcolor}
\usepackage{array}
\usepackage{colortbl}	% to color table background
\usepackage{url}

\title{\LARGE \bf
Inertial Parameter Identification \\Including Friction and Motor Dynamics
}

\author{Silvio Traversaro$^{1}$, Andrea Del Prete$^{2}$, Riccardo Muradore$^{3}$, Lorenzo Natale$^{2}$ and Francesco Nori$^{1}$% <-this % stops a space
\thanks{*This paper was supported by the FP7 EU projects CoDyCo (No. 600716 ICT 2011.2.1 Cognitive Systems and Robotics (b)), Xperience(No. 270273), and ROBOSKIN (No. 231500).}% <-this % stops a space
\thanks{$^{1}$Department of Robotics, Brains and Cognitive Sciences, Istituto Italiano di Tecnologia, Genova, Italy
        {\tt\small name.surname@iit.it}}%
\thanks{$^{2}$Department iCub Facility, Istituto Italiano di Tecnologia, Genova, Italy
        {\tt\small name.surname@iit.it}}%
\thanks{$^{3}$Department of Computer Science, University of Verona, Verona, Italy
        {\tt\small riccardo.muradore@univr.it}}%
}

\begin{document}

\maketitle
\thispagestyle{empty}
\pagestyle{empty}

%%%%%%%%%%%%%%%%%%%%%%%%%%%%%%%%%%%%%%%%%%%%%%%%%%%%%%%%%%%%%%%%%%%%%%%%%%%%%%%%
\begin{abstract}
Identification of inertial parameters is fundamental for the implementation of torque-based control in humanoids.
At the same time, good models of friction and actuator dynamics are critical for the low-level control of joint torques.
We propose a novel method to identify inertial, friction and motor parameters in a single procedure.
The identification exploits the measurements of the PWM of the DC motors and a 6-axis force/torque sensor mounted inside the kinematic chain.
The partial least-square (PLS) method is used to perform the regression.
We identified the inertial, friction and motor parameters of the right arm of the iCub humanoid robot. We verified that the identified model can accurately predict the force/torque sensor measurements and the motor voltages. Moreover, we compared the identified parameters against the CAD  parameters, in the prediction of the force/torque sensor measurements. % with a precision of ...
Finally, we showed that the estimated model can effectively detect external contacts, comparing it against a tactile-based contact detection.
The presented approach offers some advantages with respect to other state-of-the-art methods, because of its completeness (i.e. it identifies inertial, friction and motor parameters) and simplicity (only one data collection, with no particular requirements).
\end{abstract}

%%%%%%%%%%%%%%%%%%%%%%%%%%%%%%%%%%%%%%%%%%%%%%%%%%%%%%%%%%%%%%%%%%%%%%%%%%%%%%%%
%\begin{figure}[thpb]
%      \centering
%      \framebox{\parbox{3in}{We suggest that you use a text box to insert a graphic because, in an document, this method is somewhat more stable than directly inserting a picture.}}
%      %\includegraphics[scale=1.0]{figurefile}
%   \end{figure}
   
%!TEX root =  ../inertialMotorIdentification.tex
%%%%%%%%%%%%%%%%%%%%%%%%%%%%%%%%%%%%%%%%%%%%%%%%%%%%%%%%%%%%%%%%%%%%%%%%%%%%%%%%
\section{Introduction}
The relationship between joint torques and accelerations of a multi-body mechanical system is uniquely characterized by its inertial and geometric parameters.
While geometric parameters can be accurately retrieved from CAD drawings, inertial parameters (i.e. masses, centers of mass, moments of inertia) typically need to be estimated.
If we can effectively control the joint torques, inertial parameters are all we need to know to control the motion of a robot.
However, controlling joint torques is usually difficult, especially because of the large joint friction %due to the high ratio gearboxes 
that is a characteristic of gearboxes typically adopted in humanoids.

Different approaches to compensate for friction exist, but we can divide them in two big groups: feedback and feedforward.
Feedback schemes exploit torque measurements to compensate for friction, often through the design of a friction observer \cite{Tien2008} or a high-gain integral action \cite{Morel2000}.
Feedforward schemes, instead, use a previously identified friction model to predict friction, based on position, velocity and torque measurements \cite{Armstrong1988}.
Researchers proposed countless friction models \cite{Olsson1997} for control and simulation.
In this paper we consider a basic Coulomb and viscous friction model, because it captures most of the friction in our robot, while maintaining the estimation problem linear.
%In \cite{Tien2008}, Tien et al. designed a friction observer based on the motor equation and joint torque measurements; then they compensated for the observed friction. 
%This compensation scheme is equivalent to an integral term in the controller, hence its action is delayed.
%\cite{Morel2000} used a high-gain integral controller on a highly geared PUMA 550 manipulator.
%They estimated joint torques using a 6-axis force/torque sensors mounted under the manipulator.
%They reported excellent results in both torque and position control with very slow motion, but this is not surprising because integral control by nature works well for slow motion.
%They did not address the problem of medium-high velocity motion.
%We could identify inertial parameters and motor parameters separately \cite{Albu-Schaffer2001}, but this approach cannot use motor PWM measurements for the inertial parameter identification.

Dynamic parameter identification procedures are usually affected by problems such as identifiability of the considered parameters and numerical conditioning \cite{handbookident}. To solve this issues we used the  Partial Least Square (PLS) regression method \cite{geladi1986partial}.

\section{Related Works}
To localize this work in the vast literature of robot dynamics identification we need to introduce the basic equations of robot and motor dynamics.
The equation of motion of a multi-body system composed of $n$ joints and $n_B$ links can be written in a form that is linear with respect to the inertial parameters \cite{handbookident}:
\begin{equation*}
Y_{\tau}(q, \dot{q}, \ddot{q}) \phi = \tau,
\end{equation*}
where $\tau \in \mathbb{R}^n$ are the joint torques, $q \in \mathbb{R}^n$ are the joint angles, $Y_{\tau}(q, \dot{q}, \ddot{q}) \in \mathbb{R}^{n\times 10n_B}$ is the joint torque regressor matrix, and $\phi \in  \mathbb{R}^{10 n_B}$ contains the inertial parameters of all the links of the robot.
Considering DC motors as actuators, the joint torques $\tau$ are given by the difference between the motor torques $\tau_m$ and the friction torques $\tau_F$:
\begin{equation} \label{eq:tau_mdyn}
\tau = \underbrace{\Phi_m i}_{\tau_m} - \underbrace{Y_F(\dot{q}) \phi_F}_{\tau_F},
\end{equation}
where $\Phi_m = \diag{\phi_m} \in \mathbb{R}^{n\times n}$ contains the motor drive gains, $i \in \mathbb{R}^n$ are the motor currents, $\phi_F \in \mathbb{R}^{4n}$ contains four parameters for each joint describing asymmetric Coulomb and viscous frictions, and $Y_F(\dot{q}) \in \mathbb{R}^{n\times 4n}$ is the friction regressor matrix.
The same model also applies if we can control the motor voltage $v$, rather than the current $i$. 
Neglecting the electrical dynamics (which is reasonable because the dynamics of the current amplifiers have much higher bandwidth than the motors) we have that:
\begin{equation} \label{eq:mot_dyn}
i = \frac{1}{R} v + \underbrace{\frac{k_b}{R} \dot{q}}_{i_b},
\end{equation}
where $i_b$ is proportional to the back electromotive torque and $R$ is the motor coil resistance.
If we substitute \eqref{eq:mot_dyn} in \eqref{eq:tau_mdyn}, the equation maintains the same form, because the back electromotive torque can be incorporated inside the viscous friction torque, being both terms proportional to $\dot{q}$.
For this reason, in the following we do not make any distinction between voltage and current measurements.

Table~\ref{table:soa} lists different works on robot dynamics identification (including the one presented here).
%\rowcolors{2}{}{gray!15}
\begin{table}[ht] 
\caption{}
\centering 
\begin{tabular}{p{3.2cm} | p{0.3cm} p{0.3cm} p{0.3cm} p{0.3cm} | p{0.3cm} p{0.3cm} p{0.3cm}} 
 													& \multicolumn{4}{c}{\emph{Knowns}} & \multicolumn{3}{c}{\emph{Unknowns}} \\
\hline \rowcolor[gray]{.9}
      	 Author, YearÊÊÊÊÊÊÊÊÊÊÊÊÊÊÊÊÊÊÊÊ  							&    $v/i$	&   $\tau$ 	&    $w$	&    $\phi_m$	&    $\phi_m$	&    $\phi_F$	& $\phi$ 	\\ 
	 [0.5ex] \hline
	 Atkeson, 1985 \cite{An1985}							&	\x	& 		& ÊÊ		&      \x		& ÊÊÊ			& 			&ÊÊÊÊ	\x	\\ \rowcolor[gray]{.9}
	Iagnemma, 1998 \cite{Iagnemma1998}					&		& 		& ÊÊ	\x	&      		& ÊÊÊ			& 			&ÊÊÊÊ	\x	\\  
ÊÊÊÊÊÊÊÊArmstrong, 1988 \cite{Armstrong1988}					&	\x	& 		& ÊÊ		&      \x		& ÊÊÊ			& 	\x		&ÊÊÊÊ	\x	\\ \rowcolor[gray]{.9}
	Siciliano, 2009 \cite{Siciliano2009}						&	\x	& 		& ÊÊ		&      \x		& ÊÊÊ			& 	\x		&ÊÊÊÊ	\x	\\ 
	\multirow{2}*{Albu-Schaffer, 2001 \cite{Albu-Schaffer2001}}	&		& 	\x	& ÊÊ		&      		& ÊÊÊ			& 			&ÊÊÊÊ	\x	\\
													&	\x	& 	\x	& ÊÊ		&      		& ÊÊÊ	\x		& 	\x		&ÊÊÊÊ		\\ \rowcolor[gray]{.9}
	Gautier, 2011 \cite{Gautier2011}						&  \x$^*$ 	& 		& ÊÊ		&      		& ÊÊÊ	\x		& 	\x		&ÊÊÊÊ	\x	\\ 
	\multirow{2}*{Chan, 2001 \cite{Chan2001}}				&	\x	& 		& ÊÊ	\x	&      		& ÊÊÊ	\x		& 			&ÊÊÊÊ		\\ 
													&	\x	& 		& ÊÊ		&      \x		& ÊÊÊ			& 	\x		&ÊÊÊÊ	\x	\\ \rowcolor[gray]{.9}
	Traversaro, 2013 									&	\x	& 		& ÊÊ	\x	&      		& ÊÊÊ	\x		& 	\x		&ÊÊÊÊ	\x	\\ 
[0.5ex] \hline 
\end{tabular} 
\label{table:soa} %\bigskip
$^*$ They attached a known load to the manipulator.
\end{table}
\rowcolors{0}{}{}
We report which quantities are considered as \emph{known}, either because measured --- such as voltages $v$, currents $i$, joint torques $\tau$, 6-axis forces/torques $w$ --- or because known constants --- such as drive gains $\phi_m$. 
Moreover we report which parameters are estimated (i.e. motor, friction and inertial parameters).
The drive gains $\phi_m$ appear twice, because sometimes they are known, while other times they are estimated.

Atkeson et al. \cite{An1985} used current measurements and drive gains $\phi_m$ to compute motor torques $\tau_m$, and then to estimate the inertial parameters.
They could neglect friction because their manipulator was direct drive, hence it had much less friction than the highly geared robots used in all other works.
Iagnemma et al. \cite{Iagnemma1998} estimated the inertial parameters $\phi$ of a manipulator using a 6-axis F/T sensor mounted below the robot's base.
%In this case the joint friction may not be wise, because then the controller has to compensate friction by using feedback only.
Armstrong \cite{Armstrong1988} estimated motor torques $\tau_m$ from current measurements $i$ and then he identified both inertial and friction parameters.
The same approach is suggested in the well-known book by Siciliano and Sciavicco \cite{Siciliano2009}.
Albu-Schaffer and colleagues \cite{Albu-Schaffer2001} estimated $\phi$ using joint torque sensors rather than a 6-axis F/T sensor.
Moreover, they carried out a second identification to estimate motor dynamics and joint frictions.
Gautier et al. \cite{Gautier2011} used only current measurements to estimate motor, friction and inertial parameters.
In general this is possible only up to a proportional factor, because the equation to solve takes this form:
\begin{equation} \label{eq:gautier}
\mat{Y_{\tau} & -\diag{i} & Y_F} \mat{\phi \\ \phi_m \\ \phi_F} = 0
\end{equation}
Gautier solved this issue by making the robot perform the same trajectories two times, one of which with a known load attached to its end-effector.
Chan et al. \cite{Chan2001} estimated first the motor parameters $\phi_m$ using current measurements and an external F/T sensor; then they identified $\phi$ and $\phi_F$ using $i$ and the previously estimated $\phi_m$.
Our approach is similar to the work of Gautier \cite{Gautier2011}, because we estimate all parameters at the same time as they did.
However, rather than attaching a known load to the manipulator, we exploit a 6-axis F/T sensor mounted along the kinematic chain.

We now provide the reasons that led us to choose this estimation procedure.
First, even if joint torque sensors and F/T sensors allow identifying inertial parameters only, we think it is paramount to identify friction parameters too.
The final goal is to control the robot: if we do not identify friction, the controller is going to have a hard time trying to compensate for it by using feedback only.
Second, we think that it is important to identify also the drive gains $\phi_m$, even if the motor manufacturers provide them, because there may be significant inaccuracies \cite{Gautier2011}.
Moreover, the actual control signals are not the motor voltages $v$ but the PWM signals, whose proportional relationship to $v$ depends on the power supply voltage.
Since the power supply voltage may change on different platforms (and even with time), this gives us another reason to identify $\phi_m$.

%!TEX root =  ../inertialMotorIdentification.tex
%%%%%%%%%%%%%%%%%%%%%%%%%%%%%%%%%%%%%%%%%%%%%%%%%%%%%%%%%%%%%%%%%%%%%%%%%%%%%%%%%%%%%%%%%%%%%
\section{Platform}

Experiments have been conducted on the iCub \cite{Metta2010}. iCub is a full-body humanoid with 53 degrees of freedom: 6 in the head, 16 in each arm, 3 in the torso and 6 in each leg. 
%The iCub is an open-source platform and copies of the platform have been distributed among several research laboratories in Europe, America and Japan. 

Numerical computations have been implemented relying on iDyn\footnote{Doxygen documentation of the iDyn library available here: \url{http://wiki.icub.org/iCub/main/dox/html/group__iDyn.html}}, a library for computing dynamics of kinematic trees. iDyn is built on top of iKin \cite{Pattacini2010}\footnote{Doxygen documentation of the iKin library available here: \url{http://wiki.icub.org/iCub/main/dox/html/group__iKin.html}}, a library for forward-inverse kinematics of serial chains described in standard Denavit-Hartenberg notation. Both libraries are released under a GPL license.

\begin{figure} 
  \centering 
	  \includegraphics[height=1.0\hsize]{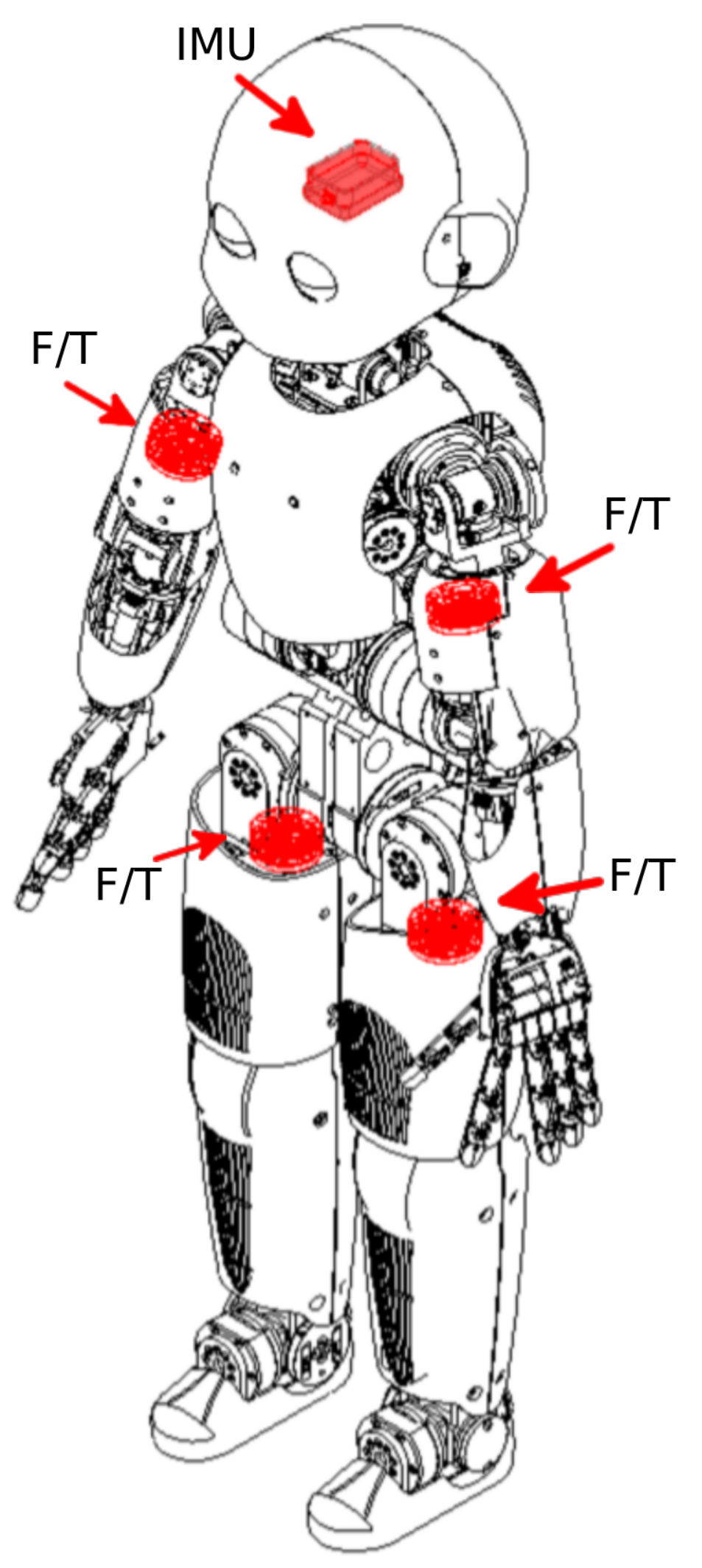} 
\caption{\label{fig:sensors} \textbf{Top}: a mechanical scheme of the humanoid robot iCub: in evidence, the four proximal six-axis F/T sensors (legs and arms) and the inertial measurement unit (IMU) (head).}
\end{figure}

As shown in Fig.~\ref{fig:sensors}, iCub is equipped with one inertial measurement unit (Xsens MTx-28A33G25\footnote{The MTx orientation tracker,\url{http://www.xsens.com/en/general/mtx}}) located in the head and with four custom-made six-axis F/T sensors, one per limb, that are placed in the middle of the limbs, as shown in Fig.~\ref{fig:sensors}.

\begin{figure} 
  \centering 
	  \includegraphics[height=0.72\hsize]{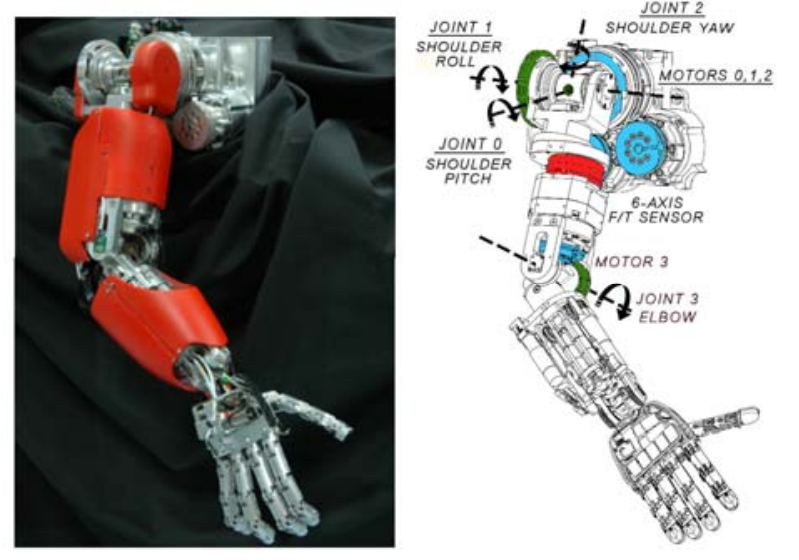} 
\caption{\label{fig:shoulder} A CAD view of the iCub arm showing its main DoFs, the location of the motors (blue) and the six-axis force/torque sensor (red).}
\end{figure}

%The iCub shoulder is a differential mechanism (see Fig.~\ref{fig:shoulder}). The robot mass distribution has been optimized housing the first three motors of the shoulder  in the upper torso. A first motor (delivering 40Nm through its 100:1 harmonic gearbox) actuates directly the first joint (shoulder pitch), whereas the other two motors (20Nm, 100:1 harmonic gearbox) actuate the remaining two joints through a set of coaxial pulleys. The ratio of the diameters of these pulleys defines a transmission reduction $r$. The resulting relationship between the motor and joint torques will be given in the following section, see in particular \eqref{eq:shoulderTau}.

\begin{figure}
\includegraphics[width=0.99\hsize]{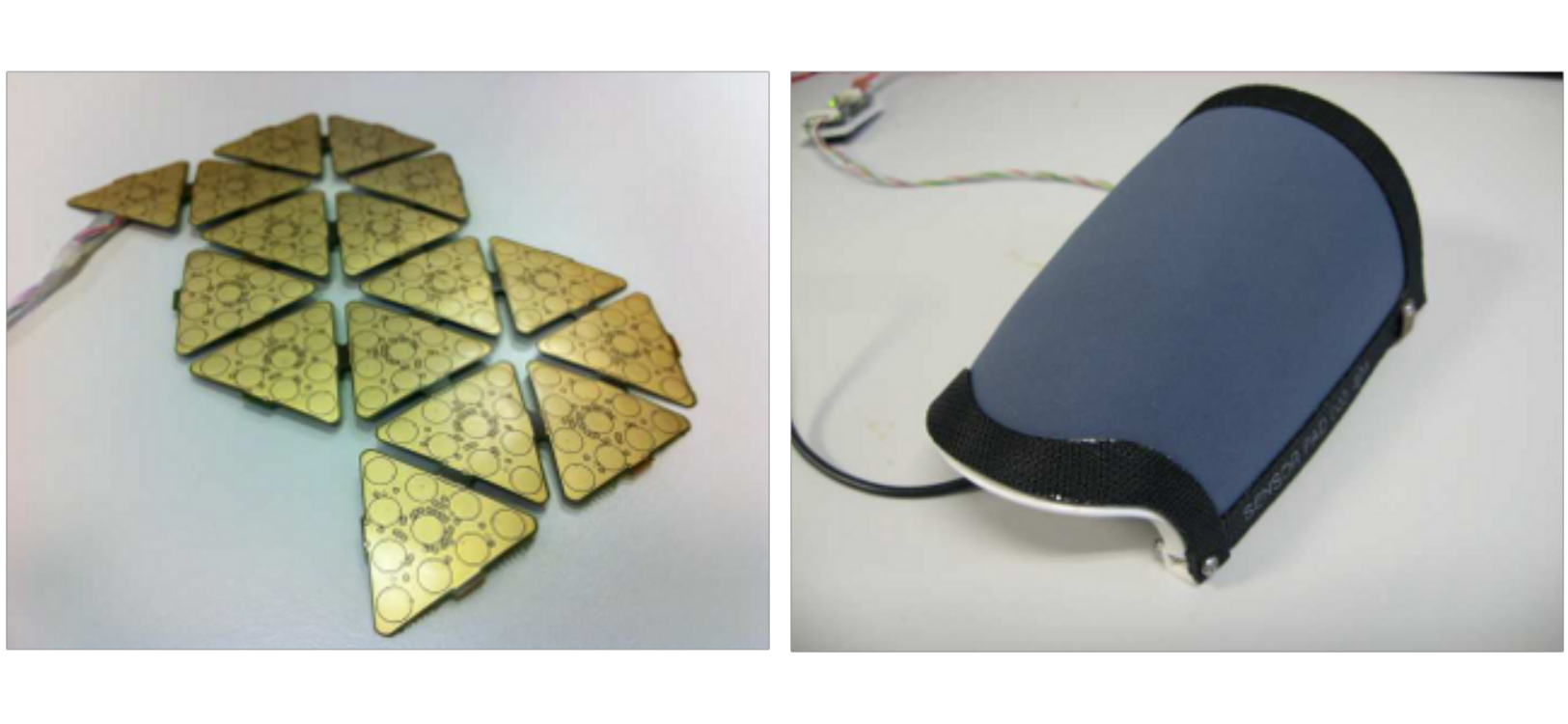}
\includegraphics[width=0.99\hsize]{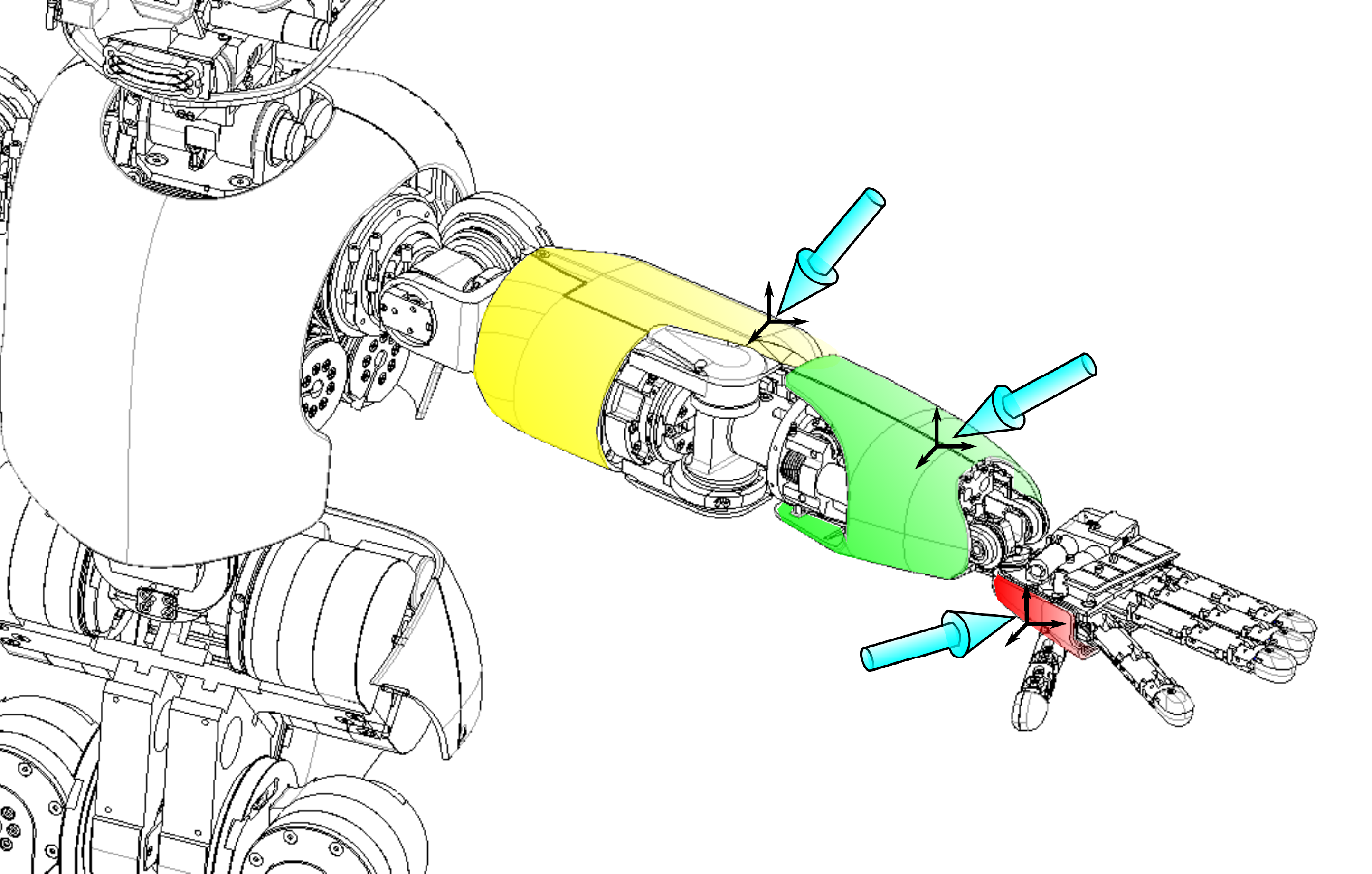}
\caption{\label{fig:skinArm}\textbf{Top}: the iCub forearm with the distributed tactile elements. The left picture shows the flexible PCB to be mounted on the forearm; the right picture shows the assembly with the capacitive element mounted on top of the lower forearm cover. Details about skin fabrication and the covering of the iCub can be found in the Roboskin website, EU Project ICT-FP7-231500 Roboskin. \textbf{Bottom}: a visualization of the iCub right arm which has been used in the experiments presented in this paper; the picture highlights with colors the sensorized areas.}
\end{figure}

The iCub upper body is covered with a distributed set of capacitive elements acting as tactile sensors \cite{Cannata2008, Maiolino2013}. At the moment the iCub upper body is covered with approximatively 2000 sensors. In this work we used these sensors to detect external contacts. The entire sensor network acts as an ``artificial skin'', constituted by a sandwich of different flexible fabrics mounted on top of a flexible Printed Circuit Board (PCB), so that the entire structure can be conformed on surfaces of different curvatures. Fig.~\ref{fig:skinArm} shows the forearm with details on the distributed tactile sensor.

%The experiments proposed in this paper have been performed on a distributed network of personal computers connected through YARP \cite{Fitzpatrick2008}, a middleware designed for exchanging data across different machines and heterogeneous operative systems. A PC104 embedded in the iCub head collects all the sensor data and it streams them on a local network through YARP ports, a software abstraction for TCP sockets. 
%Computation of the whole-body robot dynamics are performed on an external blade computers exploiting the iDyn library linked by a YARP module, which retrieves the necessary information from YARP ports. The resulting computations are then shared again on the YARP network and made available to other modules.

%!TEX root =  ../inertialMotorIdentification.tex
%%%%%%%%%%%%%%%%%%%%%%%%%%%%%%%%%%%%%%%%%%%%%%%%%%%%%%%%%%%%%%%%%%%%%%%%%%%%%%%%
\section{Method}

\subsection{Regressor structure}
%\section{Dynamics of the right arm of the iCub}
The iCub arm is composed by 6 links. The third link contains the embedded F/T sensor, so for identification purposes 
it is convenient to consider separately the inertial contribution of its two sublinks.
The inertial properties of a link are completely described by 10 parameters (mass, center of mass and inertia matrix) so the inertial 
parameters of the whole arm can be represented as a vector $\phi \in \mathbb{R}^{70}$ \cite{handbookident}.
We now examine how inertial, friction and motor parameters are related to the available measurements (i.e. internal F/T sensor and PWM of the first four motors), under the assumption that no external forces act on the arm.

\subsubsection{F/T sensor dynamics}
each of the 6 components measured by the F/T sensor has an unknown offset, which can be represented by a $6$ element vector $w_O$.
We can write the dynamics of the F/T sensor as:
\begin{equation}
\label{eq:ft_sensor_regressor}
Y_s \phi + w_O = w_s,
\end{equation}
where $w_s$ is the measured wrench, and $Y_s$ is the regressor of inertial and gravitational wrench.

\subsubsection{Elbow joint}
the elbow (joint 3) of the arm of the iCub is actuated by a dedicated motor, so we can directly write the dynamics of the joint as:
\begin{equation} \label{eq:elb_jnt}
\phi_{m_3} v_3 = Y_{\tau_3} \phi + Y_{F_3} \phi_{F_3},
\end{equation}
where:
\begin{description}
\item[$\phi_{m_k}$] is the constant relating the PWM of the motor $k$ to the exerted torque
\item[$v_k$] is the PWM of the motor $k$ 
\item[$Y_{\tau_j}$] is the regressor of inertial and gravitational torques for joint $j$
\item[$\phi_{F_j}$] is the vector of asymmetric Coulomb and viscous friction coefficients $\begin{bmatrix} \phi^{c+}_j & \phi^{c-}_j & \phi^{v+}_j & \phi^{v-}_j \end{bmatrix}^{\top}$
\item[$Y_{F_j}$] is the regressor of friction torques $\begin{bmatrix} (\sgn{\dot{q}_j})^+ & -(\sgn{\dot{q}_j})^- & (\dot{q}_j)^+ & -(\dot{q}_j)^- \end{bmatrix}$
\end{description}
The operators $()^+$ and $()^-$ select the positive and negative part (respectively) of the argument.
There is no distinction between the friction acting on the joint and on the motor, because the joint is actuated by a dedicated motor.
We can rewrite \eqref{eq:elb_jnt} as linear with respect to the unknown parameters:
\begin{equation} \label{eq:elbow_regressor}
\mat{ {Y_\tau}_3 & {Y_F}_3 & -v_3} \mat{ \phi \\ {\phi_F}_3 \\ {\phi_m}_3} = 0
\end{equation}

%%%%%%%%%%%%%%%%%%%%%%%%%%%%%%%%%%%%%%%%%%%%%%%%%%%%%%%%%%%%%%%%%%%%%%%%%%%%%%%%%%%
%%%%%%%%%%%%%%%%%%%%%%%%%%%%%%%%%%%%%%%%%%%%%%%%%%%%%%%%%%%%%%%%%%%%%%%%%%%%%%%%%%%
\subsubsection{Shoulder joints}
the shoulder group is composed of the joints 0, 1 and 2. 
We refer to this group as $I$, defining $q_I^\top =  \left[ \begin{smallmatrix} q_0 & q_1 & q_2 \end{smallmatrix} \right]$ and 
$\tau_I^\top =  \left[ \begin{smallmatrix} \tau_{0} & \tau_{1} & \tau_{2} \end{smallmatrix} \right]$.
The dynamics of the shoulder joints are:
\begin{equation} \label{eq:jnt_dyn}
\tau_{I} = Y_{\tau_I} \phi + Y_{F_I} \phi_{F_I},
\qquad
Y_{F_I} = \begin{bmatrix} Y_{F_0} & 0 & 0  \\ 0 & Y_{F_1} & 0 \\ 0 & 0 & Y_{F_2} \end{bmatrix},
\end{equation}
where: 
$$
\phi_{F_I}^{\top} = \left[ \begin{smallmatrix} 
\phi^{c+}_0 & \phi^{c-}_0 & \phi^{v+}_0 & \phi^{v-}_0 & \phi^{c+}_1 & \phi^{c-}_1 & \phi^{v+}_1 & \phi^{v-}_1  & \phi^{c+}_2 & \phi^{c-}_2 & \phi^{v+}_2 & \phi^{v-}_2 
\end{smallmatrix} \right] ^{\top}
$$
These friction coefficients are related to the joint frictions, which in this case must be distinguished from the motor frictions, because of the coupling.
The relationship between the motor torques and the joint torques is given by:
%For the shoulder group, each one of the joint torques is not related to the torque exerted by a single motor, but the torque at the motors and the torques at the joint are coupled by a matrix:
\begin{equation} \label{eq:shoulderTau}
\tau_{I} = T^{\top} \tau_{m_I}, \qquad T^{\top} = \begin{bmatrix} 1 & -r & -r \\ 0 & r & r \\ 0 & 0 & r \end{bmatrix},
\end{equation}
where $T$ is the coupling matrix, with $r = \frac{65}{40} = 1.625$. 
For further details on the shoulder coupling see \cite{parmiggiani2009}.
The relationship between the motor torques and the motor voltages is:
\begin{equation} \label{eq:sh_mot_dyn}
\tau_{m_I} = \operatorname{diag}\left(\phi_{m_I}\right)v_I - Y_{F^m_I} \phi_{F^m_I},
\end{equation}
where $Y_{F^m_I}$ is similar to the regressor $Y_{F_I}$, but with the motor velocities 
$\dot{q}_{m_I}^\top =  \left[ \begin{smallmatrix} \dot{q}_{m_0} & \dot{q}_{m_1} & \dot{q}_{m_2} \end{smallmatrix} \right]$ in place of the joint velocities $\dot{q}_I$. 
%Similarly to the torques, the joint velocities are related to the motor velocities by a coupling matrix:
%\begin{eqnarray} \label{eq:shoulderJoint}
%\dot{q}_I = T^{-1} \dot{q}_{m_I}, \qquad T^{-1} = \begin{bmatrix} 1 & 0 & 0 \\ 1 & \frac{1}{r} & 0 \\ 0 & -\frac{1}{r} & \frac{1}{r} \end{bmatrix}
%\end{eqnarray}
Substituting \eqref{eq:shoulderTau} and \eqref{eq:sh_mot_dyn} into \eqref{eq:jnt_dyn} we can write an equation that is linear with respect to all the parameters:
\begin{equation}
\label{eq:shoulder_regressor}
\mat{Y_{\tau_I} & Y_{F_I}  & T^{\top} Y_{F^m_i} & - T^{\top} \diag{v_I}} \mat{\phi \\ \phi_{F_I} \\ \phi_{F^m_I} \\ \phi_{m_I}}= 0
\end{equation}

\subsubsection{Complete regressor}
by combining all the parameters in a unique vector, we can write \eqref{eq:ft_sensor_regressor}, \eqref{eq:elbow_regressor} and \eqref{eq:shoulder_regressor} in a unified matrix equation:
\begin{equation}
\label{eq:completeRegressor}
\left[
\begin{smallmatrix} 
Y_{\tau_{I}} & 0 & Y_{F_I} & 0 & T^{\top} Y_{F^m_I} & -T^{\top} \operatorname{diag}\left(v_I\right) & 0 \\
Y_{\tau_3} & 0 & 0 & Y_{F_3} & 0 & 0 & -v_3 \\
Y_{FT} & I_{6\times6} & 0 & 0 & 0 & 0 & 0 
\end{smallmatrix}
\right]
\left[
\begin{smallmatrix}
\phi \\
w_O \\
\phi_{F_I} \\
\phi_{F_3} \\
\phi_{F_I^m} \\
\phi_{m_I} \\
\phi_{m_3} 
\end{smallmatrix}
\right]
= 
\left[
\begin{smallmatrix}
0 \\
0 \\
w_s
\end{smallmatrix}
\right]
\end{equation}
As opposed to \eqref{eq:gautier} this is a non-homogeneous equation, which allows us to avoid the ``proportional factor'' problem.
%of equation that has a single least-squares solution
%Note the difference with \eqref{eq:gautier}:  in our equation the right-hand side is not zero, which allows us to avoid the ``proportional factor'' problem.
However, not all the inertial parameters can be identified, because of the structural properties of the regressors $Y_{\tau}$ and $Y_{FT}$ \cite{handbookident}.
The regression method introduced in the next section allows us to readily solve this issue.

%!TEX root =  ../inertialMotorIdentification.tex
%%%%%%%%%%%%%%%%%%%%%%%%%%%%%%%%%%%%%%%%%%%%%%%%%%%%%%%%%%%%%%%%%%%%%%%%%%%%%%%%
\subsection{Partial Least Squares (PLS)}
\label{sec:pls}

At time $t$ equation \eqref{eq:completeRegressor} can be rewritten in a compact form as:
\begin{eqnarray}
Y_{ALL}(t)\Phi = w_{ALL}(t)
\end{eqnarray}
The identification procedure is based on time series for the matrix $Y_{ALL}(t)$ and the vector $w_{ALL}(t)$: collecting the input-output data in the matrix $A$ and the vector $b$, we end up with the linear equation:
\begin{eqnarray}
A\Phi = b + e,
\label{e:Uphi=YE}
\end{eqnarray}
where $e$ is the vector taking into account the measurement error and the unmodeled dynamics, and $\Phi$ is the unknown vector we need to estimate.

Following \cite{muradore2009statistical}, we use the Partial Least Square (PLS) method to solve (\ref{e:Uphi=YE}). This technique is very popular in chemometrics and statistical process control because it is extremely stable even with highly collinear data. The PLS goal is twofold: to explain the variance of the input matrix $A$ (as the principal component analysis, PCA, does) and to maximize the covariance between $A$ and the output vector $b$. The nonlinear iterative partial least squares (NIPALS, \cite{geladi1986partial}) computes the 
%
% is
% \begin{enumerate}
% \item Start: set $u$ equal to a column of $Y,$
% \item\label{itemNOpls} Regress columns of $X$ on $u$ to get loadings: $w^\top = \frac{u^\top X}{u^\top u},$
% \item Normalize $w$ to unit length: $w \leftarrow \frac{w}{w^\top w},$
% \item Calculate the scores: $t = \frac{X w}{w^\top w},$
% \item Regress columns of $Y$ on $t:$ $q^\top = \frac{t^\top Y}{t^\top t},$
% %\item Normalize $q$ to unit length: $q \leftarrow
% %\frac{q}{q^\top q},$
% \item Calculate new score vector for $Y$: $u = \frac{Y q}{q^\top q},$
% \item Check convergence of $u$: if YES go to \ref{itemYESpls}, if NO go to
% \ref{itemNOpls},
% \item\label{itemYESpls} Calculate $X$ matrix loadings by regressing columns of $X$ on $t$: $p^\top = \frac{t^\top X}{t^\top t},$
% %\item Regression: $b = \frac{u^\top t}{t^\top t},$
% \item Calculate residual matrices: $E=X-tp^\top$ and $F=Y-tq^\top,$
% \item To calculate the next set of latent vectors, replace $X$
% and $Y$ by $E$ and $F$ and repeat.
% \end{enumerate}
%
so-called score-loading decomposition of $A$ and $b$ as:
\begin{eqnarray}
A=\sum_{i=1}^{\nu}t_i p_i^\top + E_A = TP^\top + E_A \label{e:plsU} \\
b=\sum_{i=1}^{\nu}t_i c_i^\top + e_b = TC^\top + e_b \label{e:plsY},
\end{eqnarray}
where $\nu$ is the number of latent variables, $t_i$ are the common scores, $p_i$ and $c_i$ are the $A$- and $b$-loadings, respectively. In \cite{helland1988structure}, the structural properties of these matrices are stated. In the present case the value of $\nu$ is strictly related to the number of base parameters.

The estimation of $\Phi$ is:
\begin{eqnarray}
\label{eq:estimation_eq}
\hat\Phi = W(P^{\top}W)^{-1}C,\label{e:hat_theta_pls}
\end{eqnarray}
where $W = \left[\begin{matrix}w_1 & w_2 & \hdots & w_\nu\end{matrix}\right]$ and the vectors $w_k$ are computed in the PLS algorithm as a side-product of the score-loading decomposition. The matrix $W$ is related to $A$ by $T = A W (P^{\top}W)^{-1}$, where the matrix $P^{\top}W$ is nonsingular by construction. 

%When a new set of input-output data $(u(t),y(t))$ is available, it is possible to estimate the score $\hat t(t)$ and the output vector $\hat y(t)$. $\hat t(\cdot)$ can be seen as a set of {\it internal variables} of the system bringing complementary information w.r.t. $\hat y(t)$.

\subsection{Anomaly detection}
In process control, many indices have been developed to check if the system is {\it statistically} in- or out-of-control, mainly for safety reason (fault detection and isolation, FDI). See also \cite{muradore2012pls} for a robotic application. Among the available indices, the Hotelling or $T^2$ statistics is defined as:
\begin{eqnarray}
T^2(t) \triangleq e^\top(t)\Lambda_e^{-1}e(t), & \Lambda_e = {\rm diag}\{\sigma^2_{1},\cdots,\sigma^2_{n}\},
\label{e:T2_error}
\end{eqnarray}
where $e = y - \hat y$ is the error, and the variances $\sigma_{i}^2 = {\rm Var}\{e_i(\cdot)\}$ are computed during the identification phase. 
The plant is statistically in-control if $T^2(t) \leq T_{\alpha}^2$.
Under the assumption of normally distributed data, the upper bound $T_{\alpha}^2$ with confidence level $\alpha$ is given by:
\begin{eqnarray}
T_{\alpha}^2 \triangleq \frac{n(N-1)}{(N-n)}F_{n,N-n;\alpha},\label{e:T2limit}
\end{eqnarray}
where $n$ is the number of LVs ($\nu$ in our case), $N$ is the number of samples in the calibration data and $F$ is the Snedecor (or $F$) distribution with degrees of freedom $n$ and $N-n$, \cite{joe2003statistical}. The threshold (\ref{e:T2limit}) is only an approximation because our data is not following exactly a normal distribution and our model is an approximation of the real plant.

In this work we use the estimated and validated inertial parameters to compute $T^2$ statistics on new data, for detecting when the robot gets in contact with the environment. In other words, the ``fault'' in the present case is the contact that ``invalidates'' the estimation using $\hat\Phi$. 
%In fact, the inertial parameters have been identified in a contact-free experiment just for this purpose.

%!TEX root =  ../inertialMotorIdentification.tex
%%%%%%%%%%%%%%%%%%%%%%%%%%%%%%%%%%%%%%%%%%%%%%%%%%%%%%%%%%%%%%%%%%%%%%%%%%%%%%%%
\section{Results}
\label{sec:results}
We tested the proposed identification scheme on the right arm of the iCub.
We carried out two experiments of 5 minutes each, using impedance control to make the right hand reach pseudo-random points in the Cartesian space, without making contact with anything.
During the tests, we collected joint angle, 6-axis force/torque and PWM measurements.
Joint velocities and accelerations were estimated offline using a noncausal adaptive window technique \cite{Janabi-Sharifi2000}.
We used the first dataset to identify the model parameters and the second dataset to validate the identified model.

\begin{figure}[htbp]
\centering
\includegraphics[width=0.45\textwidth]{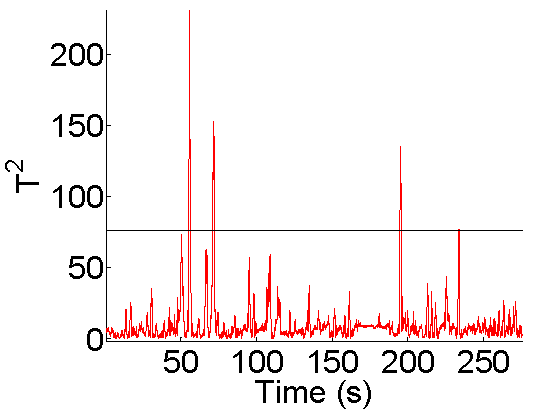}
\caption{$T^2$ index during the validation trajectory without external contacts.
The horizontal line indicates the $T_{\alpha}^2$ threshold (with $\alpha=0.99$). }
\label{fig:validation}
\end{figure}

The result of the validation are presented in Fig.~\ref{fig:validation}, which shows the validation errors using the $T^2$ index. 
On all the samples of the validation dataset, except for a few outliers, the prediction of the dynamics (using the parameters estimated on the first dataset) is lower than the confidence threshold $T_{\alpha}^2$ (with a confidence level of $\alpha=0.99$).

To further validate our model, we compared the prediction errors obtained using the parameters extracted from the 3D mechanical drawings (CAD, Computer Aided Manufacturing)\footnote{CAD models and drawings of the iCub are available under an open source license: \url{http://wiki.icub.org/wiki/RobotCub}} with those estimated using PLS. 
%These models are not perfect as they cannot be an exact description of the real robot, both for not modeled details (such as cables) or for the limited precision that every manufacturing techniques has. They are however interesting for a comparison with the estimated parameters. 
As the motor voltage measurements depend also on motor gains and friction coefficients, it is not possible to predict them using the CAD parameters,
so, for this validation, only the prediction of the F/T sensor was used. We measured a remarkable improvement in the force prediction, whereas the torque prediction error is approximately the same (see Table \ref{tab:res1}).

%    \begin{tabular}{ | c | c | c | c |}
%    \hline
%     &&$\mathbf{F}$(N)&$\boldsymbol\tau$(Nm)\\ [0.5ex] \hline
%     \multirow{2}{*}{\textit{mean}}&Estimated inertial parameters & 1.33 & 0.15 \\ 
%     &CAD inertial parameters & 1.86 & 0.15 \\ 
%     \hline
%     \multirow{2}{*}{\textit{std dev}}&Estimated inertial parameters  & 0.64 & 0.13 \\ 
%     &CAD inertial parameters  & 0.59 & 0.07 \\ [0.5ex] \hline
%    \end{tabular}
%\end{center}
%\end{table}

\begin{table}[h]
\caption{Force/torque sensor prediction errors.}
    \label{tab:res1}
\begin{center}
\begin{tabular}{p{0.9cm} | p{1.35cm} p{1.35cm}  | p{1.25cm} p{1.25cm} } 
 & \multicolumn{2}{c}{Estimated parameters} & \multicolumn{2}{c}{CAD parameters} \\ \hline
\rowcolor[gray]{.9}  &    \textit{mean}	&  \textit{std dev}	&     \textit{mean}	&    \textit{std dev}  	\\ 
[0.5ex] \hline  $\mathbf{F}$(N) &	1.33	& 	0.64	& 	1.86	&   	0.59			 \\ 
\rowcolor[gray]{.9} $\boldsymbol\tau$(Nm)	&	0.15	& 	0.13	&  0.15	&   0.07   		
\end{tabular} 
\end{center}
\end{table}

Next, we used the identified model to detect external contacts.
We carried out another experiment, similar to the previous one, but with the difference that, from time to time, we touched the iCub's arm on its tactile sensors.
Besides the measurements of the previous test, we also collected the tactile sensor measurements, to use them as ground truth for the contact detection.
To detect contacts we monitored the $T^2$ index: when it exceeds the $T_{\alpha}^2$ threshold (with $\alpha=0.99$), we infer that an external force is perturbing the system dynamics.
\begin{figure}[htbp]
\centering
\includegraphics[width=0.5\textwidth]{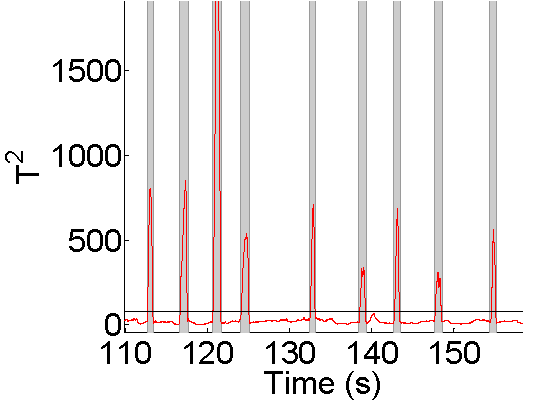}
\caption{$T^2$ index during the test with external contacts. The grey areas correspond to the periods of time when the tactile sensors were detecting contact. When the $T_{\alpha}^2$ threshold (the horizontal line) is exceeded we infer that an external force is perturbing the system dynamics.}
\label{fig:contact_detection}
\end{figure}
Fig.~\ref{fig:contact_detection} shows part of the results.
Using this threshold we obtain a true positive rate (i.e. ratio between true and actual positives) of $67\%$ and a false positive rate (i.e. ratio between false positives and actual negatives) of $6\%$.
\begin{figure}[htbp]
\centering
\includegraphics[width=0.4\textwidth]{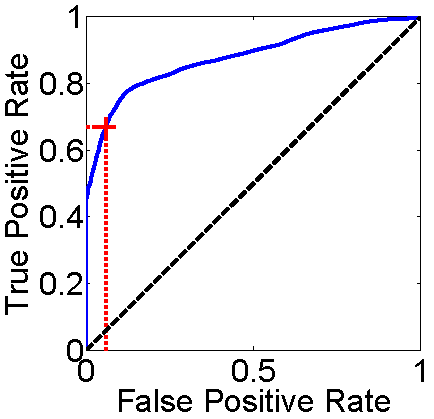}
\caption{Receiver Operating Characteristic curve for contact detection. The cross indicates the point corresponding to the $T_{\alpha}^2$ threshold, with $\alpha=0.99$.}
\label{fig:roc}
\end{figure}
Fig.~\ref{fig:roc} shows how the two rates (true positive and false positive) change in our experiment by varying the $T^2$ threshold.

%\begin{figure}[htbp]
%\centering
%\includegraphics[width=0.5\textwidth]{images/spe_skin_zoom}
%\caption{SPE index (squared prediction error) during the test with external contacts. The grey areas correspond to the periods of time when the tactile sensors were detecting contact.}
%\label{fig:res1}
%\end{figure}

\addtolength{\textheight}{-0.7cm}
%!TEX root =  ../inertialMotorIdentification.tex
%%%%%%%%%%%%%%%%%%%%%%%%%%%%%%%%%%%%%%%%%%%%%%%%%%%%%%%%%%%%%%%%%%%%%%%%%%%%%%%%
\section{Conclusions And Future Work}
\label{sec:conclusions}
In this paper we propose a new identification scheme for estimating, in a unique procedure, the inertial, friction and motor parameters of a rigid robot actuated by DC motors.
The identification exploits the measurements of the motor PWM and of an internal 6-axis force/torque sensor.
To overcome the problem of the singularity of the regressor matrix, we used the PLS regression method, which is stable also with highly collinear data.
We identified the dynamics of the right arm of the iCub robot and we verified that, in absence of external contacts, the identified model was able to accurately predict the force/torque sensor measurements.
We also proved that the error between the force/torque sensor measurement and its model-based prediction can be monitored to detect external forces.
In the experiments we used the tactile sensors of the iCub as ground truth for the contact detection task.
The key points of the presented approach are: 
\begin{itemize}
\item it estimates all the parameters that are necessary for implementing model-based controllers (e.g. computed torque control \cite{Siciliano2009})
\item it does not split the identification into subparts, so only one data collection has to be carried out \cite{Albu-Schaffer2001}
\item there is no need to attach known loads to the robot \cite{Gautier2011}
\item it is applicable to any robot having a base force/torque sensor and current/voltage measurements
\end{itemize}

We plan to use the identified parameters for implementing a model-based control architecture.
In particular, the motor and friction parameters will be used by the low-level controllers to implement joint torque control.
Inertial parameters instead will be used for computing the inverse dynamics of the robot.

Inertial parameters are time-invariant, and they cannot be estimated when the robot makes contact.
On the other hand, friction and motor parameters may change with time, so we intend to design an online estimation scheme, which can work even in presence of external contacts.
Moreover, the estimation of the drive gains would benefit from external forces; this is because, in absence of contacts, the motor torques are usually small (especially when accelerating light-weight links, such as the iCub's forearm), making the estimation of the drive gains ill-conditioned.

%%%%%%%%%%%%%%%%%%%%%%%%%%%%%%%%%%%%%%%%%%%%%%%%%%%%%%%%%%%%%%%%%%%%%%%%%%%%%%%%

%\bibliographystyle{plainnat}
{\small
\bibliographystyle{IEEEtran}
\bibliography{IEEEabrv,inertialMotorIdentification}
}

%\item The word ÒdataÓ is plural, not singular.
%\item In your paper title, if the words Òthat usesÓ can accurately replace the word ÒusingÓ, capitalize the ÒuÓ; if not, keep using lower-cased.
%\item The prefix ÒnonÓ is not a word; it should be joined to the word it modifies, usually without a hyphen.
%\item There is no period after the ÒetÓ in the Latin abbreviation Òet al.Ó.
%\item The abbreviation Òi.e.Ó means Òthat isÓ, and the abbreviation Òe.g.Ó means Òfor exampleÓ.

\end{document}